\documentclass[conference]{IEEEtran}

\usepackage{mathtools, amsmath, amssymb, amsthm}
\usepackage{tabulary}
\usepackage{tikz}
\usetikzlibrary{arrows,positioning}
\usepackage[figure,vlined,linesnumbered]{algorithm2e}
\SetAlgorithmName{Algorithm}{Algorithm}

\newcommand{\Nset}{\mathbb{N}}
\newcommand{\Rset}{\mathbb{R}}
\newcommand{\G}{\mathcal{G}}

\newcommand{\Val}{\mathit{Val}}
\newcommand{\Know}{\textsc{Know}}
\newcommand{\perm}{\textsc{Perm}}
\newcommand{\ttt}{\mathit{true}}
\newcommand{\fff}{\mathit{false}}
\newcommand{\form}{\textsc{Form}}
\newcommand{\exactly}{\textsc{exactly}}
\newcommand{\coin}{\mathit{coin}}
\newcommand{\col}{\mathit{color}}
\newcommand{\different}{\mathit{d}}
\newcommand{\peg}{\mathit{peg}}
\newcommand{\rank}{\mathit{r}}
\newcommand{\fixed}{\mathit{Fix}}
\newcommand{\dis}[1]{\langle #1 \rangle}
\newcommand{\Updates}{\mathit{Updates}}
\newcommand{\Tree}{\mathit{Tree}}
\newcommand{\stree}{\mathit{Stree}}
\newcommand{\var}{\mathit{var}}
\newcommand{\acc}{\mathit{acc}}
\newcommand{\out}{\mathit{out}}
\newcommand{\rot}{\mathit{root}}
\newcommand{\Assemble}{\textsc{Experiments}}
\newcommand{\wopt}{\textsc{Wopt}}
\newcommand{\aopt}{\textsc{Aopt}}
\newcommand{\vv}[1]{\langle #1 \rangle}

\newcommand{\Cw}{\mathcal{C}_{\mathit{worst}}}
\newcommand{\Ca}{\mathcal{C}_{\mathit{avg}}}

\tikzstyle{n}=[thick,rounded corners,draw,minimum size=1.4em,inner sep=1ex]
\tikzstyle{l}=[thick,rounded corners,draw,minimum size=1.4em,inner sep=1ex]
\tikzstyle{e}=[inner sep=.2ex,fill=white]
\tikzstyle{tran}=[thick,draw,->,>=stealth]

\newtheorem{theorem}{Theorem}
\newtheorem{example}[theorem]{Example}
\newtheorem{definition}[theorem]{Definition}

\begin{document}


\title{Strategy Synthesis for General Deductive Games Based on SAT Solving}

\author{%
\IEEEauthorblockN{Miroslav Klimo\v{s}}
\IEEEauthorblockA{\small\em Faculty of Informatics,
Masaryk University\\
Brno, Czech Republic,\\
klimos@mail.muni.cz}
\and
\IEEEauthorblockN{Anton\'{\i}n Ku\v{c}era}
\IEEEauthorblockA{\small\em Faculty of Informatics,
Masaryk University\\
Brno, Czech Republic\\
kucera@fi.muni.cz}}   


\maketitle

\begin{abstract}
  We propose a general framework for modelling and 
  solving \emph{deductive games}, where one player selects 
  a secret code and the other player strives to discover
  this code using a minimal number of allowed experiments
  that reveal some partial information about the code.
  The framework is implemented in a software tool \textsc{Cobra},
  and its functionality is demonstrated by producing 
  new results about existing deductive games.
\end{abstract}

\section{Introduction}
\label{sec-intro}

Deductive games are played by two players, the \emph{codemaker} and
the \emph{codebreaker}, where the codemaker selects a secret code from
a given finite set, and the codebreaker strives to reveal the code through
a series of \emph{experiments} whose outcomes give some partial
information about the code. A codebreaker's \emph{strategy} is a
recipe how to assemble the next experiment depending on the outcomes
of the previous experiments so that the code is eventually discovered.
The efficiency of a given strategy is measured either by the 
maximal number of experiments required to discover the code 
in the worst case, or by the expected number of experiments required 
to discover the code assuming the uniform probability distribution 
over the set of all secret codes. 

In the last decades, a lot of research has been done on special types
of deductive games such as Mastermind, ``Bulls and Cows'', the
``counterfeit coin problem (CCP)'', and others. In this paper, we use
(several variants of) Mastermind and CCP to evaluate our
results about general deductive games, and we also employ them
as running examples to illustrate various technical notions and 
abstract claims. Therefore, we introduce these games in greater detail now.

\emph{Mastermind} was invented in 1970 by Mordecai Meirowitz, an
Israeli postmaster and telecommunications expert. The codemaker
chooses a secret sequence of $n$ code pegs of $c$ colors (repetitions
allowed). The codebreaker tries to reveal the code by making guesses
(experiments) which are evaluated by a certain number of black and
white markers. A black marker is received for each code peg from the
guess which is correct in both color and position. A white marker
indicates the existence of a correct color code peg placed in the
wrong position. If there are duplicate colours in the guess, they
cannot all be awarded a marker unless they correspond to the same
number of duplicate colours in the secret code. For example, if the
code is $BACC$ and the guess is $CCAC$, then the guess is evaluated by
one black and two white markers. For the classical variant with four
pegs and six colors, Knuth \cite{Knuth:Mastermind-JRM} demonstrated a
strategy that requires five guesses in the worst case and $4.478$
guesses on average. Later, Irving \cite{Irving:Mastermind-JRM},
Neuwirth \cite{Neuwirth:Mastermind-ZOR}, and Koyama\,\&\,Lai
\cite{KL:Mastermind-JRM} presented strategies which improve the
expected number of guesses to $4.369$, $4.364$, and $4.34$,
respectively (the bound $4.34$ is already optimal).  More recently,
strategies for Mastermind were constructed semi-automatically by
using evolutionary algorithms \cite{BPR:Mastermind-evolutionary},
simulated annealing \cite{BHMOP:Mastermind-annealing}, genetic
algorithms (see, e.g., \cite{BGL:Mastermind-genetic-COR} and the
references therein), or clustering techniques
\cite{CLHH:deductivegames-EJOR}.

Interesting variants of Mastermind include ``Mastermind with
black-markers'' and ``Extended Mastermind''. The first variant, also
called ``string matching'', uses only black markers. This game was studied
already by Erd\"{o}s\,\&\,R\'{e}nyi \cite{ER:twoproblems-MTAMK} who
gave some asymptotic results about the worst-case number of
guesses. Recently, this variant found an application in genetics for
selecting a subset of genotyped individuals for phenotyping
\cite{Goodrich:MastermindAttack,GET:MastermindPhonotyping-BMC}.  The
second variant was introduced by Focardi\,\&\,Luccio in
\cite{FL:MastermindBankPins-ToCS}.  Here, a guess is not a sequence of
colors but a sequence of \emph{sets} of colors. For example, if there
are six colors and the code is $AECA$, one can make a guess
$\{A\},\{C,D,E\},\{A,B\},\{F\}$ which receives two black markers (for
the first two positions) and one white marker (for the $A$ in the
third set). It was shown in \cite{FL:MastermindBankPins-ToCS} that
this variant of Mastermind can be used to design PIN cracking
strategies for ATMs based on the so-called \emph{decimalization
  attacks}.

The basic variant of the \emph{counterfeit coin problem (CCP)} is
specified as follows. We are given $N$ coins, all identical in
appearance, and all indentical in weight except for one, which is
either heavier or lighter than the remaining $N-1$ coins. The goal is
to devise a procedure to identify the counterfeit coin using a minimal
number of weightings with a balance. This basic variant was considered
by Dyson \cite{Dyson:coins-MG} who proved that CCP can be solved 
with $w$ weightings (experiments) iff $3 \leq N \leq (3^w {-}3)/2$.
There are numerous modifications and generalizations of the basic 
variant (higher number of counterfeit coins, additional regular coins,
multi-pan balance scale, parallel weighting, etc.) which are harder
to analyze and in some cases only partial results exist. We refer to
\cite{GN:coins-overview-AMM} for an overview. 
\smallskip

\noindent
\textbf{Our contribution:} In this paper, we propose a generic model
for deductive games based on propositional logic (see
Section~\ref{sec-model}), and we design a general algorithm for
synthesizing efficient codebreaker's strategies (see
Section~\ref{sec-equiv}). When assembling the next experiment
performed by the constructed strategy, our synthesis algorithm first
eliminates ``equivalent'' experiments to avoid the state-space
explosion. We design strategy synthesis algorithms both for
\emph{ranking} strategies, which try to identify the ``most
promising'' experiment using a given ranking function, and for
\emph{optimal} strategies, where the worst or average number of
experiments is minimized.  The whole framework is
implemented in a software tool \textsc{Cobra}.  Some new results about
existing deductive games achieved with this tool are presented in
Section~\ref{sec-exp}. To the best of our knowledge, this is
the first attempt for establishing a unified framework for
modelling and analyzing \emph{general} deductive games without
focusing on some particular class of instances. 





\section{Preliminaries}
\label{sec-prelim}

The set of all positive integers is denoted by $\Nset$.
For a given set $\Sigma$, we use $|\Sigma|$ to denote the cardinality
of $\Sigma$, and $\Sigma^*$ to denote the set of all finite sequences
(words) over $\Sigma$. In particular, $\varepsilon \in \Sigma^*$ denotes
the empty word. For a given $k \in \Nset$, the set of all
\mbox{$k$-tuples} of elements in~$\Sigma$ is denoted by 
$\Sigma^k$. The $i$-th component of 
$\vec{p} \in \Sigma^k$ is denoted by $\vec{p}_i$ for all
$1 \leq i \leq k$. We also use $\Sigma^{\dis{k}}$ to denote the
subset of $\Sigma^k$ consisting of all $\vec{p} \in \Sigma^k$
such that the components of $\vec{p}$ are pairwise different.
Given $\vec{p} \in \Sigma^k$ and $\vec{q} \in \Sigma^m$, we write
$\vec{p}\vec{q}$ to denote the tuple $\vec{r} \in \Sigma^{k+m}$
where $\vec{r}_i = \vec{p}_i$ for all $1 \leq i \leq k$ and 
$\vec{r}_{k+i} =\vec{q}_i$ for all $1 \leq i \leq m$. We also
write $\vec{p}[i/a]$ to denote the tuple which is the same as
$\vec{p}$ except that $\vec{p}[i/a]_i = a$.
The set of all total functions from $\Sigma$ to $X$, where $X$ 
is a set, is denoted by $X^{\Sigma}$.

We assume familiarity with basic notions of propositional logic.
Given a set $A$, the set of all propositional formulae over $A$ 
is denoted by $\form(A)$. Apart of standard Boolean connectives, 
we also use the operator $\exactly_i$, where $i \in \Nset$, such that 
$\exactly_i(\varphi_1,\ldots,\varphi_m)$ is true iff exactly $i$ of 
the formulae $\varphi_1,\ldots,\varphi_m$ are true.
For technical convenience, we assume that \emph{all}
Boolean connectives used in formulae of $\form(A)$ are commutative.
That is, we allow for $\neg,\wedge,\vee,\exactly_i,\ldots$, but we
forbid implication which must be expressed using the allowed operators.
For a given formula $\varphi \in \form(A)$, we use
$\Val(\varphi)$ to denote the set of all valuations of $A$
satisfying $\varphi$. We write $\varphi \approx \psi$ and 
$\varphi \equiv \psi$ to denote that $\varphi$ and $\psi$ are
semantically and syntactically equivalent, respectively, and we
extend this notation also to sets of formulae. Hence, if $\Phi,\Psi$
are sets of formulae, then $\Phi \approx \Psi$ and $\Phi \equiv \Psi$
means that the two sets are the same up to the respective equivalence.
The syntactic equivalence $\equiv$ is considered modulo basic
identities such as commutativity or associativity.




\section{A Formal Model of Deductive Games}
\label{sec-model}

In this section we present a generic mathematical model for
deductive games based on propositional logic. Intuitively, 
a deductive game is specified by 
\begin{itemize}
\item a finite set $X$ of propositional variables and a propositional
  formula $\varphi_0$ over $X$ such that every secret code $c$ can be
  represented by a unique valuation $v_c$ of $X$, and for every valuation
  $v$ of $X$ we have that $v(\varphi_0) = \ttt$ iff $v = v_c$ for some
  secret code $c$;
\item a finite set of allowed experiments $T$. 
\end{itemize}
To model CCP with $N$ coins, we put $X = \{x_1,\ldots,x_N,y\}$, 
and we represent a secret code $c$ where the $i$-th coin is lighter/heavier
by a valuation $v_c$ where $v_c(x_i) = \ttt$, $v_c(x_j) = \fff$ for all
$j \neq i$, and $v_c(y) = \ttt$ (i.e., $y$ is set to $\ttt$ iff 
the different coin is heavier). The formula $\varphi_0$ says that 
precisely one of the variables $x_1,\ldots,x_N$ is set to~$\ttt$.
In Mastermind with $n$ pegs and $m$ colors, the set $X$ contains
variables $x_{i,j}$ for all $1 \leq i \leq n$ and $1 \leq j \leq m$;
the variable $x_{i,j}$ is set to $\ttt$ iff the $i$-th peg has 
color~$j$. The formula $\varphi_0$ says that each peg has precisely
one color.

Typically, the number of possible experiments
is large but many of them differ only in the concrete choice of 
participating objects. For example, in CCP with $6$ coins there are 
essentially three types of experiments (we can weight 
either $1+1$, $2+2$, or $3+3$ coins) which are instantiated by a
concrete selection of coins. 
In Mastermind, we perform essentially only one type of experiment 
(a guess) which is instantiated by a concrete tuple of colors. In
general, we use a finite set $\Sigma$ of \emph{parameters} to represent
the objects (such as coins and colors) participating in experiments.
A \emph{parameterized experiment} $t \in T$ is a triple $(k,P,\Phi)$ where $k$
is the number of parameters, $P \subseteq \Sigma^k$ is the 
set of admissible instances, and $\Phi$ are possible outcomes. 
In CCP, all parameters (coins) must be pairwise different, so 
$P = \Sigma^{\dis{k}}$. In Mastermind, the  parameters (colors) used in a guess 
can be freely repeated, so $P = \Sigma^n$ where $n$ is the number 
of pegs. Possible  outcomes of~$t$ are given as 
\emph{abstract propositional formulae} (see below). Now we state 
a formal definition of a deductive game.

\begin{definition}
\label{def-game}
  A \emph{deductive game} is a tuple $\G = (X,\varphi_0,\Sigma,F,T)$, where
  \begin{itemize} 
  \item $X$ is a finite set of propositional variables, 
  \item $\varphi_0 \in \form(X)$ is a satisfiable \emph{initial constraint}, 
  \item $\Sigma$ is a finite set of \emph{parameters},
  \item $F \subseteq X^\Sigma$ is a set of \emph{attributes}  
    such that for all $f,f' \in F$ 
    where $f \neq f'$ we have that the images of $f$ and $f'$ are disjoint,   
  \item $T$ is a finite set of \emph{parameterized experiments} of the form
    $(k,P,\Phi)$ where $k \in \Nset$ is the number of parameters,
    $P \subseteq \Sigma^k$ is a set of \emph{instances}, 
    and 
    \mbox{$\Phi \subseteq \form(X \cup \{f(\$j) \mid f \in F, 
    1 {\leq} j {\leq} k\})$} is
    a finite set of \emph{outcomes}.
  \end{itemize}
\end{definition}

\noindent
The intuition behind $X$, $\varphi_0$, and $\Sigma$ is explained
above. Each attribute $f \in F$ corresponds to some 
``property'' that every object $a \in \Sigma$ either does or
does not satisfy, and $f(a)$ is the propositional variable of $X$ which 
encodes the $f$-property of~$a$. In CCP with $N$ coins, for each
object (coin) we need to encode the property of ``being
different''. So, there is just one attribute $\different$ which maps
$\coin_i$ to $x_i$ for all $1 \leq i \leq N$.  In Mastermind with $n$
pegs and $m$ colors, each object (color) has the property of ``being
the color of peg~$i$'', where $i$ ranges from $1$ to~$n$.  Hence, there are $n$
attributes $\peg_1,\ldots,\peg_n$ where $\peg_i(\col_j) = x_{i,j}$.

Now consider a parameterized experiment $t = (k,P,\Phi)$. An
\emph{instance} of $t$ is a $k$-tuple $\vec{p} \in P \subseteq \Sigma^k$ 
of parameters.
For every instance $\vec{p} \in P$ and every outcome $\psi \in \Phi$, we
define the \emph{\mbox{$\vec{p}$-instance} of $\psi$} as the formula
$\psi(\vec{p}) \in \form(X)$ obtained from $\psi$ by substituting
each atom $f(\$j)$ with the variable $f(\vec{p}_j)$. Hence, $f(\$j)$
denotes the variable which encodes the \mbox{$f$-attribute} of
$\vec{p}_j$. In the rest of this paper, we typically use $\varphi,\psi$ to 
range over outcomes, and $\xi,\chi$ to range over their instances.



\begin{example}
\label{exa-coins}
  CCP with four coins can be modeled as a deductive game 
  $\G = (X,\varphi_0,\Sigma,F,T)$ where 
  \begin{itemize}
  \item $X = \{x_1,x_2,x_3,x_4,y\}$,
  \item $\varphi_0 = \exactly_1(x_1,x_2,x_3,x_4)$,
  \item $\Sigma = \{\coin_1,\coin_2,\coin_3,\coin_4\}$,
  \item $F = \{\different\}$ where $\different(\coin_i) = x_i$ 
     for every $1 \leq i \leq 4$,
  \item $T = \{t_1,t_2\}$ where
    \begin{itemize}
    \item[] $t_1  =  (2,\Sigma^{\dis{2}},\{\varphi_<,\varphi_=,\varphi_> \})$
    \item[] $t_2  =  (4,\Sigma^{\dis{4}},\{\psi_<,\psi_=,\psi_> \})$
    \end{itemize}
    and 
    \begin{itemize}
    \item[] \makebox[1.5em][l]{$\varphi_<$} 
          $ = \ (\different(\$1) \wedge \neg y) \,\vee\, (\different(\$2) \wedge y)$
    \item[] \makebox[1.5em][l]{$\varphi_=$}
          $ = \ \neg \different(\$1)\, \wedge \,\neg \different(\$2)$
    \item[] \makebox[1.5em][l]{$\varphi_>$}
          $ = \ (\different(\$1) \wedge y) \,\vee\, (\different(\$2) \wedge \neg y)$
    \item[] \makebox[1.5em][l]{$\psi_<$}
          $ = \ ((\different(\$1) \vee \different(\$2)) \wedge \neg y) \,\vee\, 
                        ((\different(\$3) \vee \different(\$4) \wedge y)$
    \item[] \makebox[1.5em][l]{$\psi_=$}
          $ = \ \neg \different(\$1)\, \wedge \,\neg \different(\$2)\, \wedge 
                       \,\neg \different(\$3)\, \wedge \,\neg \different(\$4)$
    \item[] \makebox[1.5em][l]{$\psi_>$}
          $ = \ ((\different(\$1) \vee \different(\$2) )\wedge y) \,\vee\, 
                       ((\different(\$3) \vee \different(\$4) ) \wedge \neg y)$
    \end{itemize}
  \end{itemize}
  Note that $t_1$ and $t_2$ correspond to weightings of $1+1$ and $2+2$
  coins, respectively. 
  The formulae $\varphi_<$, $\varphi_=$, and $\varphi_>$ encode the
  three possible outcomes of weighting $1+1$ coins. In particular,
  $\varphi_<$ describes the outcome when the left pan is lighter; then
  we learn that either the first coin is different and lighter, or
  the second coin is different and heavier. If we put 
  $\vec{p} = (\coin_4,\coin_3)$, then $\varphi_<(\vec{p})$ is
  the formula $(x_4 \wedge \neg y) \,\vee\, (x_3 \wedge y)$.
\end{example}

In the following, we also use $E$ to denote the set of all
\emph{experiment instances} (or just \emph{experiments})
defined by 
\[
   E = \{(t,\vec{p}) \mid t \in T,\ \vec{p} \mbox{ is an instance of } t \}.
\]

Note that Definition~\ref{def-game} does not impose any restrictions
on the structure of parameterized experiments. In general, the 
knowledge accumulated by performing experiments 
may even become inconsistent. Obviously, it makes no sense to  
``solve'' such wrongly specified games. In our next definition we 
introduce a subset of \emph{well-formed} deductive games where 
no consistency problems arise. Intuitively, we require that for each
valuation of $\Val(\varphi_0)$, every experiment produces
exactly one valid outcome.

\begin{definition}
\label{def-well-formed}
We say that a deductive game $\G = (X,\varphi_0,\Sigma,F,T)$ is
\emph{well-formed} if for every $v \in \Val(\varphi_0)$ 
and every experiment $(t,\vec{p}) \in E$
there is exactly one outcome $\psi$ of $t$ such that
$v(\psi(\vec{p})) = \ttt$.
\end{definition}

\noindent
Deductive games that correctly encode meaningful problems
(such as the game of Example~\ref{exa-coins}) are well-formed,
so this condition is not restrictive.
Still, our tool \textsc{Cobra} (see Section~\ref{sec-exp}) verifies that
the game on input is well-formed by invoking an optimized algorithm 
which only considers a subset of experiments which represents $E$
up to a suitable ``experiment equivalence'' (see Section~\ref{sec-equiv}).



\section{Solving Deductive Games}
\label{sec-solving}

Now we introduce the notion of codebreaker's strategy,
explain what we mean by solving a deductive game, and then define
some special types of strategies that are important for purposes
of automatic strategy synthesis.

For the rest of this section, we fix a well-formed deductive game $\G
= (X,\varphi_0,\Sigma,F,T)$. For every experiment $e = (t,\vec{p})$, 
we use $\Phi(e)$ to denote the set of \mbox{$\vec{p}$-instances} 
of all outcomes of~$t$. An \emph{evaluated experiment} is a pair 
$(e,\xi)$, where $\xi \in \Phi(e)$. The set of all evaluated 
experiments is denoted by $\Omega$. 

Intuitively, the game $\G$ is played as follows:
\begin{itemize}
\item[1.] The codemaker selects a secret code $v \in \Val(\varphi_0)$.
\item[2.] The codebreaker selects the next experiment $e \in E$.
\item[3.] The codemaker evaluates the experiment $e$ against $v$
  and returns the resulting evaluated experiment $(e,\xi)$.
\item[4.] If the codemaker has enough information to determine~$v$, 
  the play ends. Otherwise, it continues with Step~2.
\end{itemize}
We assume that the only information available to the codebreaker
is the history of evaluated experiments played so far. This is reflected
in the next definition.

\begin{definition}
\label{def-strategy}
  A \emph{strategy} is a (total) function $\sigma : \Omega^* \rightarrow E$
  which specifies the next experiment for a given finite history of
  evaluated experiments.
\end{definition}

Every strategy $\sigma$ determines the associated \emph{decision tree},
denoted by $\Tree_\sigma$, where the internal nodes are labelled by 
experiments, the leaves are labeled by valuations of $\Val(\varphi_0)$,
and the edges are labeled by evaluated experiments. For every node 
$u$ of $\Tree_\sigma$, let 
$
  \lambda_u^\sigma =  (e_1,\xi_1),\dots,(e_n,\xi_n)
$
be the unique sequence of evaluated experiments that label the edges
of the unique finite path from the root of $\Tree_\sigma$ to~$u$ (note that
if $u$ is the root, then $\lambda_u^\sigma = \varepsilon$). We also use
$\Psi_u^\sigma$ to denote the formula
$
   \varphi_0 \wedge \xi_1 \wedge \cdots \wedge \xi_n 
$. 
The structure of $\Tree_\sigma$ is completely determined by the following 
conditions that must be satisfied by $\Tree_\sigma$: 
\begin{itemize}
\item For every node $u$ of $\Tree_\sigma$, the label of $u$ is either 
  $\sigma(\lambda_u^\sigma)$ or the only valuation of $\Val(\Psi_u^\sigma)$, 
  depending on whether \mbox{$|\Val(\Psi_u^\sigma)| > 1$} or not, respectively. 
\item Every node $u$ of $\Tree_\sigma$ labeled by $e \in E$ has a unique
  successor $u_\xi$ for each $\xi \in \Phi(e)$ such that
  the formula $\Psi_u^\sigma \wedge \xi$ is still satisfiable. The edge from $u$ 
  to $u_\xi$ is labeled by $(e,\xi)$.
\end{itemize}
Note that different nodes/edges may have the same labels, and
$\Tree_\sigma$ may contain infinite paths in general.

\begin{figure}[ht]
\begin{center}
\begin{tikzpicture}[x=1.2cm,y=1.5cm,font=\scriptsize]
  \node (E1)  at (0,0)   [n] {$e_1 =(t_1,(1,2))$};    
  \node (E2l) at (-2,-1) [n] {$e_2 =(t_1,(1,3))$};    
  \node (E2c) at (0,-1)  [n] {$e_2 =(t_1,(1,3))$};    
  \node (E3)  at (2,-1)  [n] {$e_3 =(t_1,(2,4))$};    
  \node (E4)  at (0,-3)  [n] {$e_4 =(t_1,(1,4))$};    
  \node (L1)  at (-3,-2) [l] {$1,\ell$};
  \node (L2)  at (-1.5,-2) [l] {$2,h$};
  \node (L3)  at (-2,-3) [l] {$3,h$};
  \node (L4)  at (2,-3)  [l] {$3,\ell$};
  \node (L5)  at (1.5,-2)  [l] {$2,\ell$};
  \node (L6)  at (3,-2)  [l] {$1,h$};
  \node (L7)  at (-1,-4)  [l] {$4,h$};
  \node (L8)  at (1,-4)   [l] {$4,\ell$};
  \draw [tran,rounded corners] (E1) -- +(-2,-.3) -- node[e] 
     {$(e_1,\varphi_{<}(1,2))$} (E2l);
  \draw [tran,rounded corners] (E1) -- +(0,-.3) -- node[e] 
     {$(e_1,\varphi_{=}(1,2))$} (E2c);
  \draw [tran,rounded corners] (E1) -- +(2,-.3) -- node[e] 
     {$(e_1,\varphi_{>}(1,2))$} (E3);
  \draw [tran,rounded corners] (E2l) -- +(.5,-.3) -- node[e] 
     {$(e_2,\varphi_{=}(1,3))$} (L2);
  \draw [tran,rounded corners] (E2l) -- +(-1,-.3) -- node[e] 
     {$(e_2,\varphi_{<}(1,3))$} (L1);
  \draw [tran,rounded corners] (E3) -- +(-.5,-.3) -- node[e] 
     {$(e_3,\varphi_{<}(2,4))$} (L5);
  \draw [tran,rounded corners] (E3) -- +(1,-.3) -- node[e] 
     {$(e_3,\varphi_{=}(2,4))$} (L6);
  \draw [tran,rounded corners] (E2c) -- +(-1,-1.3) -- +(-2,-1.3) --  node[e] 
     {$(e_2,\varphi_{<}(1,3))$} (L3);
  \draw [tran,rounded corners] (E2c) -- +(1,-1.3) -- +(2,-1.3) --  node[e] 
     {$(e_2,\varphi_{>}(1,3))$} (L4);
  \draw [tran,rounded corners] (E2c) -- +(0,-1.3) -- +(0,-1.3) --  node[e] 
     {$(e_2,\varphi_{=}(1,3))$} (E4);
  \draw [tran,rounded corners] (E4)  -- +(-1,-.3) -- node[e] 
     {$(e_4,\varphi_{<}(1,4))$} (L7);
  \draw [tran,rounded corners] (E4)  -- +(1,-.3) -- node[e] 
     {$(e_4,\varphi_{>}(1,4))$} (L8);
\end{tikzpicture}
\end{center}
\caption{A decision tree for a simple strategy.}
\label{fig-tree}
\end{figure}
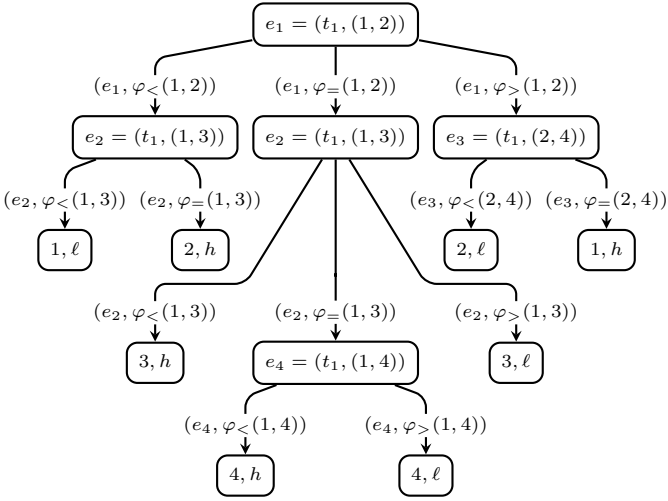

\begin{example}
\label{exa-tree}
Consider the game $\G$ of Example~\ref{exa-coins}. A decision tree for
a simple strategy $\sigma$ is shown in Fig.~\ref{fig-tree} 
(we write just $i$ instead
of $\coin_i$, and we use $i,\ell$ (or $i,h$) to denote the valuation
of $\Val(\varphi_0)$ which sets $x_i$ to $\ttt$ and $y$ to $\fff$ 
(or to $\ttt$, respectively). Note that $\sigma$ discovers the secret
code by performing at most three experiments. Also note that some internal
nodes have only two successors, because the third outcome is impossible. 
\end{example}

Since $\G$ is well-formed, every strategy $\sigma$ and every 
$v \in \Val(\varphi_0)$ determine a unique (finite or infinite) 
path $u_1,u_2,u_3,\ldots$ initiated in the root of $\Tree_\sigma$,
which intuitively correspond to a \emph{play} of $\G$ where the codemaker
selects the secret code~$v$. We use 
\[
  \lambda_{v}^\sigma = (e_1,\xi_1),(e_2,\xi_2),(e_3,\xi_3),\ldots 
\]
to denote the associated sequence of evaluated experiments (i.e.,
$(e_i,\xi_i)$ is the label of $(u_i,u_{i+1})$; we also use
$\lambda_{v}^\sigma(i)$ to denote the $i$-th evaluated experiment of
$\lambda_{v}^\sigma$). The length of $\lambda_{v}^\sigma$ is denoted by
$\#\lambda_{v}^\sigma$.
Further, for every $k \leq \#\lambda_{v}^\sigma$, 
we use $\Psi_{v}^\sigma[k]$ to denote 
the formula $\Psi_{u_k}^\sigma$ which represents the knowledge
accumulated after evaluating the first $k$~experiments.
%

Now we can also define the 
\emph{worst/average case complexity of~$\sigma$}, denoted by
$\Cw(\sigma)$ and $\Ca(\sigma)$, in the following way:
\begin{align*}
  \Cw(\sigma) & = 
    \max \{ \#\lambda_{v}^\sigma \mid v \in \Val(\varphi_0)\}\\
  \Ca(\sigma) & = 
    \frac{\sum_{v \in \Val(\varphi_0)} \#\lambda_{v}^\sigma}{|\Val(\varphi_0)|}
\end{align*}
Note that the worst/average case complexity of $\sigma$ is finite iff
\emph{every} $v \in \Val(\varphi_0)$ is discovered by $\sigma$ after a finite 
number of experiments. 

\begin{definition}
\label{def-optimal}
  We say that $\G$ is \emph{solvable} iff there exists a strategy 
  $\sigma$ with a finite worst/average case complexity.
  
  Further, we say that a strategy $\sigma$ is \emph{worst case optimal} iff for 
  every strategy $\sigma'$ we have that $\Cw(\sigma) \leq \Cw(\sigma')$.
  Similarly, $\sigma$ is \emph{average case optimal} iff
  $\Ca(\sigma) \leq \Ca(\sigma')$ for every strategy $\sigma'$.
\end{definition}

\noindent
For example, the strategy~$\sigma$ of Example~\ref{exa-tree} is worst case
optimal (cf.{} the lower bound of Dyson \cite{Dyson:coins-MG} mentioned
in Section~\ref{sec-intro}).

In general, a codebreaker's strategy may depend not only on the outcomes
of previously evaluated experiments, but also on their order. Now we
show that the codebreaker can actually ``ignore'' all aspects 
of a play except for the accumulated knowledge.

\begin{definition}
\label{def-knowledge}
  A strategy $\sigma$ is \emph{knowledge-based} if for all 
  $v_1,v_2 \in \Val(\varphi_0)$ and $k \in \Nset$ such that
  $\Psi_{v_1}^\sigma[k] \approx \Psi_{v_2}^\sigma[k]$ we have that 
  $
     \sigma(\lambda_{v_1}^\sigma(1),\ldots,\lambda_{v_1}^\sigma(k)) \ = \
     \sigma(\lambda_{v_2}^\sigma(1),\ldots,\lambda_{v_2}^\sigma(k)) 
  $.
\end{definition}

\noindent
Observe that a knowledge-based strategy depends only on 
the \emph{semantics} of accumulated knowledge in a play.

Let $\Know \subseteq \form(X)$ be the set of all formulae 
representing an accumulated knowledge, i.e.,
$\Know$ consists of all $\Psi_{v}^\sigma[k]$ where 
$\sigma$ is a strategy, $v \in \Val(\varphi_0)$, and $k \in \Nset$.
Every knowledge-based strategy $\sigma$ can then be equivalently 
defined as a function $\tau : \Know \rightarrow E$ where
$\tau(\Psi_{v}^\sigma[k]) = \sigma(\lambda_{v}^\sigma(1),\ldots,
\lambda_{v}^\sigma(k))$, and for all
equivalent $\varphi_1,\varphi_2 \in \Know$ we have that
$\tau(\varphi_1) = \tau(\varphi_2)$. In the rest of this paper,
we adopt this alternative definition,
and we use $\tau$ to range over knowledge-based strategies. 

The next theorem says that knowledge-based strategies are equally powerful
as general strategies. 
 
\begin{theorem}
\label{thm-knowledge}
  Let $\G$ be a well-formed deductive game. For every strategy $\sigma$
  there exists a knowledge-based strategy $\tau$ such that for every
  $v \in \Val(\varphi_0)$ we have that 
  $\#\lambda_{v}^\tau \leq \#\lambda_{v}^\sigma$.
\end{theorem}

\noindent
Consequently, for every well-formed deductive game there exist 
worst/average case optimal strategies that are knowledge-based.

For purposes of automatic strategy synthesis, abstract
knowledge-based strategies are not sufficiently workable.
Intuitively, a knowledge-based strategy somehow ``ranks'' the
outcomes of available experiments and tries to identify
the most promising experiment which decreases the ``uncertainty'' 
of the accumulated knowledge as much as possible. The notion 
of ``ranking'' is not explicitly captured in 
Definition~\ref{def-knowledge}. Therefore, we also introduce
\emph{ranking} strategies, which are equally powerful as 
knowledge-based strategies, but reflect the above intuition 
explicitly.

For every accumulated knowledge $\varphi \in \Know$ and every experiment
$e \in E$, we define the set 
\[
 \Updates[\varphi,e]= \{\varphi \wedge \xi \mid \xi \in \Phi(e)\}
\]
which represents possible ``updates'' in the accumulated 
knowledge that can be obtained by performing~$e$. Every
experiment $e$ is then ranked by a fixed \emph{ranking function} 
\mbox{$r : 2^{\Know} \rightarrow \Rset$} which is applied to
the set $\Updates[\varphi,e]$. The corresponding \mbox{$r$-ranking} strategy 
selects an experiment with the minimal rank; if there are several 
candidates, some fixed auxiliary total ordering $\preceq$ over $E$ 
is used, and the least candidate w.r.t.{}~$\preceq$ is selected.  

\begin{definition}
\label{def-ranking}
  Let $\rank : 2^{\Know} \rightarrow \Rset$ be a \emph{ranking function},
  and $\preceq$ a total ordering over the set $E$ of all experiments.
  A \emph{ranking strategy determined by $\rank$ and 
  $\preceq$} is a function 
  \mbox{$\tau[\rank,\preceq] : \Know \rightarrow E$}
  such that $\tau[\rank,\preceq](\varphi)$ is the least
  element of $\{ e\in E \mid \Updates[\varphi,e] = \mathit{Min}\}$ 
  w.r.t.{}~$\preceq$,
  where $\mathit{Min} = \min \{\Updates[\varphi,e'] \mid e' \in E\}$. 
\end{definition}

\noindent
For every knowledge-based strategy $\tau$ there is an ``equivalent''
ranking strategy $\tau[\rank,\preceq]$ where, for all 
$\varphi \in \Know$ and $e \in E$, the value of 
$r(\Updates[\varphi,e])$ is either $0$ or $1$, depending
on whether $\Updates[\varphi,e]$ is equal to 
$\Updates[\varphi,\tau(\varphi)]$ or not, respectively. The ordering
$\preceq$ can be chosen arbitrarily. One can easily show that 
for every $v \in \Val(\varphi_0)$ we have that 
$\#\lambda_{v}^\tau = \#\lambda_{v}^{\tau[r,\preceq]}$. So, ranking 
strategies are equally powerful as knowledge-based strategies 
and hence also general strategies. In particular, there exist
worst/average case optimal ranking strategies, but it is not 
clear what kind of ranking functions they need to employ.

Now we introduce several distinguished ranking functions. 
They generalize concepts previously used for solving Mastermind,
and there are also two new rankings based on the number of fixed
variables. The associated ranking strategies
always use the lexicographical ordering over $E$
determined by some fixed linear orderings over the sets $T$ 
and~$\Sigma$.
\begin{itemize}
  \item $\textbf{max-models}(\Psi) = \max_{\psi \in \Psi} |\Val(\psi)|$.
     The associated ranking strategy minimizes the worst-case number 
     of remaining secret codes. For Mastermind, this was suggested
     by Knuth~\cite{Knuth:Mastermind-JRM}.
  \item $\textbf{exp-models}(\Psi) = 
     \frac{\sum_{\psi \in \Psi} |\Val(\psi)|^2}{\sum_{\psi \in \Psi} |\Val(\psi)|}$.
     The associated ranking strategy minimizes the expected number 
     of remaining secret codes. For Mastermind, this was suggested
     by Irwing~\cite{Irving:Mastermind-JRM}.
  \item $\textbf{ent-models}(\Psi) = 
     \sum_{\psi \in \Psi} \frac{|\Val(\psi)|}{N} \cdot 
     \log(\frac{|\Val(\psi)|}{N})$, where $N = \sum_{\psi \in \Psi}|\Val(\psi)|$.
     The associated ranking strategy minimizes the entropy of the numbers 
     of remaining secret codes. For Mastermind, this was suggested
     by Neuwirth~\cite{Neuwirth:Mastermind-ZOR}.
  \item $\textbf{parts}(\Psi) = -|\{\psi \in \Psi \mid \psi \mbox{ is 
     satisfiable} \}|$. The associated ranking strategy minimizes the
     number of satisfiable outcomes. For Mastermind, this was suggested
     by Kooi~\cite{Kooi:Mastermind-ICGA}.
\end{itemize}
We say that a variable $x \in X$ is \emph{fixed} in a formula $\varphi
\in \form(X)$ if $x$ is set to the same value by all valuations
satisfying $\varphi$ (i.e., for all $v,v' \in \Val(\varphi)$ we have
that $v(x) = v'(x)$). The set of all variables that are fixed in
$\varphi$ is denoted by $\fixed(\varphi)$. We consider two ranking
functions based on the number of fixed variables.
\begin{itemize}
\item $\textbf{min-fixed}(\Psi) = 
    - \min_{\psi \in \Psi} |\fixed(\psi)|$. The associated ranking function 
    maximizes the number of fixed variables.
\item $\textbf{exp-fixed}(\Psi) =  
    - \frac{\sum_{\psi \in \Psi} |\Val(\psi)|\cdot|\fixed(\psi)|}%
    {\sum_{\psi \in \Psi}|\Val(\psi)|}$. The associated ranking function 
    maximizes the expected number of fixed variables.
\end{itemize}
Intuitively, a ``good'' ranking function should satisfy two requirements:
\begin{itemize}
\item The associateted ranking strategy should have a low worst/average
  case complexity (see Definition~\ref{def-optimal}). Ideally, it
  should be optimal.
\item The ranking function should be easy to evaluate for a given
  experiment~$e$. This is crucial for automatic strategy synthesis.
\end{itemize}
Obviously, there is a conflict in these two requirement. For example,
the $\textbf{max-models}$ ranking often produces a rather efficient
strategy, but the number of satisfying valuations of a given
propositional formula is hard to compute. On the other hand, 
$\textbf{min-fixed}$ ranking produces a good ranking strategy
only in some cases (e.g., for CCP and its variants), but it is relatively
easy to compute with modern SAT solvers even for large formulae.
We explain these issues in greater detail in the next two~sections,
where we also provide some experimental results.

\section{Equivalent Experiments, Strategy Synthesis}
\label{sec-equiv}


Intuitively, one of the main problems we have to tackle when computing
a good strategy for solving~$\G$ is the large number of experiments.
For example, in CCP with $60$ coins, there are more than
$10^{63}$ ways of instantiating the parameterized experiment 
$t$ formalizing the weighting of $20 + 20$ coins.
If $t$ is performed in the first step, i.e., when the accumulated 
knowledge is just $\varphi_0$, then \emph{all} instances of $t$ 
are ``equivalent'' in the sense that the knowledge learned by these 
instances is the same up to some symmetry (i.e., a permutation 
of coins). Hence, it suffices to consider only \emph{one} instance 
of $t$ and disregard the others. In this section, we present a general 
algorithm which, for a given accumulated knowledge~$\varphi$, 
computes a subset of experiments $S_\varphi$ such that every experiment
$e \in E$ is \mbox{``$\varphi$-equivalent''} to some experiment of~$S_\varphi$.
This algorithm substantially improves the efficiency of the whole framework.

For the rest of this section, we fix a well-formed 
deductive game $\G = (X,\varphi_0,\Sigma,F,T)$.
A \emph{permutation} of $X$ is a bijection $\pi : X \rightarrow X$.
We use $\perm(X)$ to denote the set of all permutations of~$X$. Given a 
formula $\varphi \in \form(X)$ and a permutation $\pi \in \perm(X)$, 
we use $\pi(\varphi)$ to denote the formula obtained from $\varphi$
by simultaneously substituting every occurrence of every $x \in X$ with 
$\pi(x)$. For a given $\Phi \subseteq \form(X)$, we use
$\pi(\Phi)$ to denote the set $\{\pi(\varphi) \mid \varphi \in \Phi\}$.
 
\begin{definition}
\label{def-equivalence}
  Let $e,e' \in E$ and $\pi \in \perm(X)$. We say that $e'$ is 
  \emph{$\pi$-symmetrical} to $e$ if 
  $\pi(\Phi(e)) \approx \Phi(e')$.  
  A \emph{symmetry group of $\G$}, denoted by $\Pi$, consist of
  all $\pi \in \perm(X)$ such that for every $e \in E$ there is a 
  $\pi$-symmetrical $e' \in E$. 

  We say that $e,e' \in E$ are \emph{equivalent w.r.t.{} a given 
  $\varphi\in\Know$}, written $e \sim_\varphi e'$, if there is
  $\pi \in \Pi$ such that
  \[
    \{\varphi \wedge \psi \mid \psi \in \Phi(e)\} \approx
    \{\pi(\varphi \wedge \varrho) \mid \varrho \in \Phi(e')\}\, .
  \]
\end{definition}

\noindent
Note that $\Pi$ is indeed a group, i.e., $\Pi$ contains the identity 
and if $\pi \in \Pi$, then the inverse $\pi^{-1}$ of $\pi$
also belongs to~$\Pi$.

\begin{example}
  Consider the game $\G$ of Example~\ref{exa-coins}. Then
  $\Pi = \{\pi \in \perm(X) \mid \pi(y) = y\}$. Hence, for
  all $\vec{p},\vec{q} \in \Sigma^{\dis{4}}$ we have that
  $(t_2,\vec{p}) \sim_{\varphi_0} (t_2,\vec{q})$, and the partition
  $E/{\sim_{\varphi_0}}$ has only two equivalence classes corresponding
  to $t_1$ and $t_2$. For $\varphi = \varphi_0 \wedge \neg(x_1 \vee x_2)$,
  we have that 
  $(t_1(\coin_1,\coin_2)) \sim_{\varphi} (t_2,(\coin_3,\coin_1,\coin_2,\coin_4))$.
\end{example}

As we shall see, it usually suffices to consider only one experiment for 
each equivalence class of $E/{\sim_{\varphi}}$, which improves
the efficiency of our strategy synthesis algorithms presented
at the end of this section. These algorithms invoke
a function $\Assemble(\varphi)$ parameterized by $\varphi\in\Know$ 
which computes a set of experiments 
$S_\varphi \subseteq E$ such that for every $e \in E$ there is at least
one $e' \in S_\varphi$ where $e \sim_\varphi e'$. 

A naive approach to constructing $S_\varphi$ is to initialize 
$\hat{S}_\varphi := \emptyset$ and then process every $t = (k,P,\Phi) \in T$
as follows: for every $\vec{p} \in \Sigma^k$, we check whether
$\vec{p} \in P$ and $(t,\vec{p}) \not\sim_\varphi e$ for all 
$e \in \hat{S}_\varphi$; if this test is positive, we 
put $\hat{S}_\varphi := \hat{S}_\varphi \cup \{(t,\vec{p})\}$, and continue
with the next~$\vec{p}$. When we are done with all $t \in T$, we set
$S_\varphi := \hat{S}_\varphi$.
Obviously, this trivial algorithm is inefficient for at least two reasons.
\begin{itemize}
\item[1.] The size of $\Sigma^k$ can be very large (think of CCP with
  $60$ coins), and it may not be possible to go over all $\vec{p}\in\Sigma^k$.
\item[2.] The problem of checking $\sim_\varphi$ is computationally 
  hard. 
\end{itemize}
Now we show how to overcome these issues. Intuitively, the first issue
is tackled by optimizing the trivial backtracting algorithm which would
normally generate all elements of $\Sigma^k$ lexicographically
using some total ordering $\preceq$ over $\Sigma$. We improve the
functionality of this algorithm as follows: when the backtracking
algorithm is done with generating all $k$-tuples starting with a given 
prefix $\vec{u}a \in \Sigma^m$, where $m \in \{1,\ldots,k\}$, and aims 
to generate all $k$-tuples starting with $\vec{u}b$ where
$b$ is the $\preceq$-successor of~$a$, we first check whether
$b$ is \emph{dominated} by $a$ for $\varphi$, $t$, and $\vec{u}$.
If it is the case, we continue immediately with 
the \mbox{$\preceq$-successor} of~$b$, i.e., we do \emph{not} examine
the \mbox{$k$-tuples} starting with $\vec{u}b$ at all (note that 
the \mbox{$\preceq$-successor} of~$b$ is again checked for 
dominance~by~$a$). The dominance by~$a$ can be verified quickly and guarantees
that all of the ignored experiments are equivalent to some of the 
already generated ones. As we shall see, 
this can lead to drastic improvements in the total
number of generated instances which can be \emph{much} smaller 
than $|\Sigma|^k$. The set of all experiments generated in the first 
phase is denoted by $S_\varphi^1$.
 
The second issue is tackled by designing an algorithm which tries to
decide $\sim_\varphi$ for a given pair of experiments $e_1,e_2$ by
first removing the \emph{fixed variables} (see Section~\ref{sec-solving}) 
in $\varphi$ and the outcomes of
$e_1,e_2$ using a SAT solver, and then constructing two labeled
graphs $B_{\varphi,e_1}$ and $B_{\varphi,e_1}$ which are checked for
isomorphism (here we again rely on existing software tools). If the graphs are
isomorphic, we have that $e_1 \sim_\varphi e_2$, and we can safely
remove $e_1$ or $e_2$ from $S_\varphi^1$. When the
experiments are ordered by some $\preceq$, we prefer to remove the larger
one.  Thus, we produce the set~$S_\varphi$.  Now we explain
both phases in greater detail.

Let $t= (k,P,\Phi) \in T$. For all $i \in \{1,\ldots,k\}$, let
$F_i$ be the set of all $f \in F$ such that some $\psi \in \Phi$
contains the atom $f(\$i)$. Positions $i,j \in \{1,\ldots,k\}$
are \emph{compatible} if $i \neq j$ and $F_i \cap F_j \neq \emptyset$.
Further, we define the set $X_t = \{f(\vec{p}_i) \mid f \in F_i, 
\vec{p} \in P, 1 \leq i \leq k\}$.
We say that $t$ is \emph{faithful} if it satisfies the following conditions:
\begin{itemize}
\item The variables of $X_t$ do not occur in any $\psi \in \Phi$.
\item For all compatible $i,j$ and $\vec{p} \in P$ we have
   that $\vec{p}_i \neq \vec{p}_j$.
\item For all compatible $i,j$ and $\vec{p} \in P$ we have
   that $\vec{p}[i/\vec{p}_j,j/\vec{p}_i] \in P$.
\item For all $\vec{p} \in P$, $i \in \{1,\ldots,k\}$, and $b \in \Sigma$
   such that $b \neq \vec{p}_j$ for every $j$
   compatible with $i$ we have that $\vec{p}[i/b] \in P$.
\end{itemize}
One can easily verify that all experiments in the game of 
Example~\ref{exa-coins} are faithful, and the same holds for 
the game formalizing Mastermind. Note that faithfulness
is particularly easy to verify if $P = \Sigma^k$ or $P = \Sigma^{\dis{k}}$.

Let us assume that $t = (k,P,\Phi) \in T$ is faithful. We say that
$\vec{r} \in \Sigma^i$, where $1 \leq i \leq k$, is
\emph{\mbox{$t$-feasible}} if there is $\vec{s} \in \Sigma^{k-i}$
such that $\vec{r}\vec{s} \in P$. 
Now, let us fix some $m \in \{1,\ldots,k\}$, $\vec{u} \in \Sigma^{m-1}$, and
$a\in \Sigma$ such that the \mbox{$m$-tuple} $\vec{u}a$ is 
\mbox{$t$-feasible}. Further, let $\vec{q} = \vec{u}b\vec{v} \in P$ for some
$b \in \Sigma$ and $\vec{v} \in \Sigma^{k-m}$. Then
there exists at most one~$j$ compatible with~$m$ such that 
$\vec{q}_j = a$ (if there were two such indexes $j,\ell$, we 
could ``swap'' $\vec{q}_m$ and $\vec{q}_j$ in $\vec{q}$ and thus 
obtain an instance of $P$ which does not satisfy the second 
condition of faithfulness).  
If there is no such~$j$, we put $\vec{p} = \vec{q}[m/a]$ and 
$\hat{F} = F_m$. Otherwise, we have that $j>m$ 
(if $j<m$, then $\vec{u}a$ is not \mbox{$t$-feasible}), 
and we put $\vec{p} = \vec{q}[m/a,j/b]$ and $\hat{F} = F_i \cup F_j$. 
Observe that $\vec{p} \in P$ because $t$ is faithful.
We also define the associated permutation $\hat{\pi} \in
\perm_X$, where $\hat{\pi}(f(a)) = f(b)$ and $\hat{\pi}(f(b)) = f(a)$
for all $f \in \hat{F}$, and $\hat{\pi}(y) = y$ for the other
variables. For $e = (t,\vec{q})$, we use $\hat{e}$ to denote 
the associated experiment $(t,\vec{p})$. The underlying 
$m$, $\vec{u}$, $\vec{v}$, $a$ and $b$ are always clearly 
determined by the context.


\begin{definition}
\label{def-inter}
  Let $\varphi\in\Know$, $t = (k,P,\Phi) \in T$, and  
  $\vec{u}a \in \Sigma^{m}$ a $t$-feasible tuple, where $1 \leq m <k$. 
  We say that $b \in \Sigma$ is \emph{dominated} by $a$ 
  for $\varphi$, $t$, and $\vec{u}$, if either $\vec{u}b$ is not 
  \mbox{$t$-feasible},
  or $\vec{u}b$ is \mbox{$t$-feasible}, $t$ is faithful, and the 
  following condition is satisfied:
  \begin{itemize}
    \item for all experiments of the form $e = (t,\vec{u} b \vec{v})$ 
    we have that $\hat{\pi} \in \Pi$ and
    $\{\varphi \wedge \psi \mid \psi \in \Phi(e)\} \equiv
       \{\hat{\pi}(\varphi \wedge \varrho) \mid \varrho 
           \in \Phi(\hat{e})\}$. 
  
  \end{itemize}
\end{definition}

\noindent
Note that the last condition of Definition~\ref{def-inter} guarantees
that $e \sim_\varphi \hat{e}$ (cf. Definition~\ref{def-equivalence}).
Also observe that $\hat{e} \preceq e$. Hence, the
requirement that $b$ is dominated by $a$ for $\varphi$, $t$, and $\vec{u}$
fully justifies the correctness of the improved backtracking algorithm
discussed above in the sense that the resulting set $S_\varphi^1$
indeed contains at least one representative for each equivalence class
of $E/{\sim_{\varphi}}$. Also observe that in the last condition of
Definition~\ref{def-inter}, we use the \emph{syntactic} equality of
two sets of propositional variables, which is easy to check. Further,
we do not need to consider \emph{all} $\vec{v} \in \Sigma^{k-m-1}$
when verifying this condition; the only important information about
$\vec{v}$ is whether $\vec{v}$ contains $a$ at a position compatible 
with~$m$. Hence, we need to examine $k-m$ possibilities in the worst case.
Checking whether $\hat{\pi} \in \Pi$ is not trivial in general, and
our tool Cobra handles only some restricted cases (e.g., when all 
experiments allow for arbitrary or no parameter repetition). 

Now we describe the second phase, when we try to identify and remove
some equivalent experiments in~$S_\varphi^1$. The method works only under
the condition that for every  $t = (k,P,\Phi) \in T$ we have that
$P$ is closed under all permutations of~$\Sigma$ (note that this condition
is satisfied when $P = \Sigma^k$ or $P = \Sigma^{\dis{k}}$). Possible
generalizations are left for future work. The method starts by 
constructing a labeled \emph{base graph} $B = (V,E,L)$ of $\G$, where
the set of vertices $V$ is $X \cup F$ (we assume $X \cap F = \emptyset$)
and the edges of $E$ are determined as follows: 
\begin{itemize}
\item $(f,x) \in E$, where $f \in F$ and $x \in X$, if there is $a \in \Sigma$
   such that $f(a) = x$;
\item $(x,y) \in E$, where $x,y \in X$, if there are $a \in \Sigma$,
   \mbox{$f,g \in F$},  $t \in T$, some outcome $\psi$ of $T$, such that 
   $f(a) = x$, $g(a) =y$, and both $f(\$i)$ and $g(\$i)$ appear in $\psi$
   for some  $i \in \{1,\ldots,k\}$.
\end{itemize} 
The labelling $L : V \rightarrow X \cup F \cup \{\var\}$, where 
$\var \not\in X \cup F$, assigns $\var$ to every variable $x \in X$ such 
that $x$ does not appear in any outcome of any parameterized experiment
of~$T$. For the other vertices $v \in V$, we have that $L(v) = v$. 
The base graph $B$ represents a subset of $\Pi$ in the following sense:

\begin{theorem}
\label{thm-auto}
  Let $\pi$ be an automorphism of~$B$. Then $\pi$ restricted to $X$ is
  an element of~$\Pi$.
\end{theorem}
 
Theorem~\ref{thm-auto} is proven by constructing a $\pi$-symmetrical
experiment to a given parameterized experiment~$(t,\vec{p})$. 
Now, let $\varphi \in \form_X$ be a formula representing the accumulated 
knowledge, and let $e_1 = (t_1,\vec{p})$ and $e_2 = (t_2,\vec{q})$
be experiments. We show how to construct two labeled graphs $B_{\varphi,e_1}$
and $B_{\varphi,e_2}$ such that the existence of an isomorphism between
$B_{\varphi,e_1}$ and $B_{\varphi,e_2}$ implies $e_1 \sim_\varphi e_2$.

For every formula $\psi \in \form_X$, let $\stree(\psi)$ be the syntax
tree of $\psi$, where every inner node is labeled by the associated
Boolean operator, the leaves are labeled by the associated variables
of~$X$, and the root is a fresh vertex $\rot(\psi)$ with only one
successor which corresponds to the topmost operator of $\psi$ (the
label of $\rot(\psi)$ is irrelevant for now). Recall that we only
allow for commutative operators, so the ordering of successors of a
given inner node of $\stree(\psi)$ is not significant.  Each such
$\stree(\psi)$ can be \emph{attached} to any graph $B'$ which subsumes
$B$ by taking the disjoint union of the vertices of $B'$ and the inner
vertices of $\stree(\psi)$, and identifying all leaves of
$\stree(\psi)$ labeled by $x \in X$ with the unique node $x$ of~$B'$.
All edges and labels are preserved.

The graph $B_{\varphi,e_1}$ is obtained by subsequently attaching
$\stree(\overline{\varphi}),\stree(\overline{\psi_1(\vec{p})}),\ldots,
\stree(\overline{\psi_n(\vec{p})})$ to the base graph of~$B$, where 
$\psi_1,\ldots,\psi_n$ are the outcomes of~$t_1$, and for every
$\psi \in \form(X)$, the formula $\overline{\psi}$ is obtained from~$\psi$ 
by removing its \emph{fixed variables} (see Section~\ref{sec-solving})
using a SAT solver. The root of $\stree(\overline{\varphi})$ is labelled
by $\acc$, and the roots of $\stree(\overline{\psi_1(\vec{p})}),\ldots,
\stree(\overline{\psi_n(\vec{p})})$ 
are labeled by~$\out$. The graph $B_{\varphi,e_2}$ is constructed in the same
way, again using the labels $\acc$ and~$\out$.  

\begin{theorem}
\label{thm-iso}
  If $B_{\varphi,e_1}$, $B_{\varphi,e_2}$ are isomorphic, then 
  \mbox{$e_1 \sim_\varphi e_2$}.
\end{theorem}

Intuitively, an isomorphism between $B_{\varphi,e_1}$ and $B_{\varphi,e_2}$
encodes a permutation $\pi \in \Pi$ (see Theorem~\ref{thm-auto})
which witnesses the equivalence of $e_1$ and $e_2$ w.r.t.{} $\varphi$.

The procedure $\Assemble(\varphi)$ is used to compute decision trees
for ranking strategies and optimal worst/average case strategies
in the following way.
Let $\tau[r,\preceq]$ be a ranking strategy such that for all
$e_1,e_2 \in E$ and $\varphi \in \Know$ we have that $e_1 \sim_\varphi e_2$
implies $r(e_1) = r(e_2)$. Note that all ranking functions introduced
in Section~\ref{sec-solving} satisfy this property. The decision tree
$\Tree_{\tau[r,\preceq]}$ is computed top-down. When we need to determine 
the label of a given node $u$ where the associated accumulated knowledge
is $\Psi_u$, we first check whether $|\Val(\Psi_u)| = 1$ using a
SAT solver. If it is the case, we label $u$ with the only valuation
of $\Val(\Psi_u)$. Otherwise, we need to compute the experiment
$\tau[r,\preceq](\Psi_u)$ (see Definition~\ref{def-ranking}). 
It follows immediately that $\tau[r,\preceq](\Psi_u)$ is contained in
$S_{\Psi_u} := \Assemble(\Psi_u)$. Hence, we label $u$ with the least element 
of $\{ e\in S_{\Psi_u} \mid \Updates[\Psi_u,e] = \mathit{Min}\}$ 
w.r.t.{}~$\preceq$, where 
$\mathit{Min} = \min \{\Updates[\Psi_u,e'] \mid e' \in S_{\Psi_u}\}$.
This element is computed with the help of a SAT solver.  

The way of computing a decision tree for an optimal worst/average case
strategy is more involved. Let $\wopt_{\G}$ and $\aopt_{\G}$ be the sets
of all knowledge-based strategies which are worst case optimal and
average case optimal, respectively. First, observe that if 
$\tau \in \wopt_{\G}$ and $\tau(\varphi) = e$ for some 
$\varphi \in \Know$, then for every $e' \in E$ where $e \sim_\varphi e'$
there is $\tau' \in \wopt_{\G}$ such that $\tau'(\varphi) = e'$. Hence,
we can safely restrict the range of $\tau(\varphi)$ to 
$\Assemble(\varphi)$. Further, if $\tau(\varphi) = e$ and 
$\varphi' \equiv \pi(\varphi)$ for some $\pi \in \Pi$, we can safely
put $\tau(\varphi') = \pi(e)$. The same properties hold also for
the strategies of $\aopt_{\G}$. 

A recursive function for computing 
a worst/average case optimal strategy is show in Fig.~\ref{alg:acopt}.
The function is parameterized by $\varphi \in \Know$ and an upper bound
on the worst/average number of experiments performed by an optimal
strategy for the initial knowledge $\varphi$. The function returns
a pair $\vv{e_\varphi,C_\varphi}$ where $e_\varphi$ is the experiment
selected for $\varphi$ and $C_\varphi$ is the worst/average
number of experiments that are needed to solve the game for the
initial knowledge~$\varphi$. Hence, the algorithm
is invoked by $\textsc{Optimal}(\varphi_0,\infty)$.
Note that the algorithm caches the computed results
and when it encounters that $\varphi$ is \mbox{$\pi$-symmetric}
 to some previously
processed formula, it uses the cached results immediately (line~3).
The lines executed only when constructing the worst (or average) 
case optimal strategy are prefixed by $[W]$ (or $[A]$, respectively).   
At line~4, the constant $\mathit{Out}$ is equal to 
$\max_{(k,P,\Phi) \in T} |\Phi(t)|$. Obviously, we need at least
$\lceil \log_{\mathit{Out}}(|\Val(\varphi)|) \rceil$ experiments to distinguish
among the remaining $|\Val(\varphi)|$ alternatives.

\begin{algorithm}[t]\footnotesize
\DontPrintSemicolon
\SetKwProg{optfun}{Function}{}{}
\optfun{\textsc{Optimal}\textnormal{($\varphi$, $\mathit{upper}$)}}{
\lIf{$|\Val(\varphi)| = 1$}{\Return{$\vv{v,0}$} where $v \in \Val(\varphi)$}
\lIf{$\varphi$ is cached }{\Return{ the cached result}}
[W] \lIf{$\lceil \log_{\mathit{Out}}(|\Val(\varphi)|) \rceil > \textit{upper}$}%
    {\Return{$\vv{\mathit{err},\infty}$}}
  $S_\varphi := \Assemble(\varphi)$\;
$\mathit{best} := \mathit{upper};\ e_\varphi := 
   \textit{some element of } S_\varphi$\;
\For{$e \in S_\varphi$}
{
  $ val := 0 $\;
  \For{$\psi\in\Phi(e)$} {
    \If{$\mathit{SAT}{(\varphi\wedge\psi)}$}{
    $\vv{e_\psi,C_\psi} := 
         \textsc{Optimal}(\varphi \wedge \psi,\mathit{best}-1)$\;
[W] $val := \max( val, 1 + C_\psi)$\;
[A] $val := val + |\Val(\varphi \wedge\psi)| \cdot (1+C_\psi)$\;
    }
  }
  [A] $val := val \;/\; |\Val(\varphi)|$\;
  \lIf{$val \leq \mathit{best}$}{$\mathit{best} := val; e_\varphi := e$}
}
Cache the result $\vv{e_{\varphi},\mathit{best}}$ for $\varphi$\;
\Return{$\vv{e_{\varphi},\mathit{best}}$}
}{}
\caption{\small Computing optimal strategies.}
\label{alg:acopt}
\end{algorithm}

\section{Experimental Results}
\label{sec-exp}

The framework for modelling and analyzing deductive games described in
previous sections has been implemented in our software tool
\textsc{Cobra}\footnote{The tool is freely available at
  \texttt{https://github.com/myreg/cobra}.}. In this section we
present selected experimental results which aim to demonstrate the
efficiency of the algorithm for eliminating symmetric experiments, and
to show that the framework is powerful enough to produce new results
about existing deductive games and their variants. In all these experiments,
we employ the SAT solver MiniSat \cite{ES:MiniSat} and the tool 
Bliss \cite{JK:Bliss} for checking graph isomorphism.

The functionality of $\Assemble(\varphi)$ can be well demonstrated 
on CCP and Mastermind. Consider CCP with $26$, $39$, and $50$ coins.
The next table shows the \emph{average} size of $S_\varphi^1$ and $S_\varphi$
when computing the $i$-th experiment in the decision tree for
\textbf{max-models} ranking strategy (see Section~\ref{sec-solving}).
The total number of experiments for $26$, $39$ and $50$ coins is 
larger than $10^{16}$, $10^{27}$, and $10^{39}$, respectively. 
Observe that for $26$ 
and $39$ coins, only four experiments are needed to reveal the 
counterfeit coin, and hence the last row is empty.
\bigskip

\noindent
{\scriptsize
\begin{tabulary}{.5\textwidth}{|R|RR|RR|RR|}
\hline
  & \multicolumn{2}{c|}{\textbf{CCP 26} \raisebox{2.1ex}{~}} &
    \multicolumn{2}{c|}{\textbf{CCP 39}} &
    \multicolumn{2}{c|}{\textbf{CCP 50}} \\
 &  \multicolumn{2}{c|}{($\approx 10^{16}$ exp.)} &
    \multicolumn{2}{c|}{($\approx 10^{27}$ exp.)} &
    \multicolumn{2}{c|}{($\approx 10^{39}$ exp.)} \\\hline

Exp.No.\rule{0pt}{1em} & Phase~1 & Phase~2 & Phase~1 & Phase~2 & Phase~1 & Phase~2 \\\hline
1 & 13.0 & 13.0 &  19.0 & 19.0 & 25.0 & 25.0 \\
2 & 4,365.0 & 861.7  & 26,638.7 & 3,318.0  & 83,625.0 & 8,591.0 \\
3 & 603.0 & 36.4  & 2,263.0 & 88.1 & 5,733.4 & 172.2  \\
4 & 76.3 & 4.2 & 214.7 & 7.2 & 405.1 & 10.4  \\
5 & - & - & - & - & 153.2 & 4.1  \\\hline
  \end{tabulary}}  
\bigskip

\noindent
Note that in the first round, \emph{all} equivalent 
experiments are discovered already in the first phase, i.e., when
computing~$S_1$. These experiments correspond to the number of coins
that can be weighted (e.g., for $50$ coins we can weight 
$1{+}1, \dots, 25{+}25$ coins, which gives $25$ experiments).
In the second round, when we run $\Assemble(\varphi)$ for three 
different formulae $\varphi \in \Know$, the average size of $S^1_\varphi$ 
is already larger, and the second phase (eliminating equivalent experiments)
further reduces the average size of the resulting $S_\varphi$. 

A similar table for Mastermind is shown below. Here we consider 
three variants with $3/8$, $4/6$, and $5/3$ pegs/colors. The table
shows the average size of $S_\varphi$ when computing the $i$-th experiment
in the decision trees for \textbf{max-models} and \textbf{parts} 
ranking strategies. 
\bigskip

\noindent
{\scriptsize%
\begin{tabulary}{.5\textwidth}{|R|RR|RR|RR|}
\hline
 & \multicolumn{2}{c|}{\textbf{MM 3x8} (512 exp.)} &
   \multicolumn{2}{c|}{\textbf{MM 4x6} (1296 exp.)} &
   \multicolumn{2}{c|}{\textbf{MM 5x3} (243 exp.)} \\
Exp.No. & max-mod. & parts & max-mod. & parts & max-mod. & parts\\\hline
1 & 3.00 & 3.00 & 5.00 & 5.00 & 5.00 & 5.00 \\
2 & 17.38 & 17.38 & 34.91 & 106.62 & 59.25 & 59.25 \\
3 & 72.31 & 87.83 & 243.40 & 580.03 & 121.45 & 186.90 \\
4 & 71.54 & 87.98 & 344.02 & 417.02 & -  & - \\
5 & 25.36 & 31.97 & - & - & - & - \\\hline
\end{tabulary}}
\bigskip

\noindent
Note that for Mastermind, the reduction is more efficient for 
more colors and less pegs, and that the values for the two ranking
strategies significantly differ, which means that they divide the solution 
space in a rather different way.

Now we present examples of results obtained by running our tool that
(to the best of our knowledge) have not yet been published in the
existing literature about deductive games. 

The first example concerns 
CCP. While the worst case complexity of CCP is fully understood 
\cite{Dyson:coins-MG}, we are not aware of any results about 
the \emph{average} case complexity of CPP. Using \textsc{Cobra}, we were 
able to compute the \emph{average-case optimal strategy} for up to 
$60$~coins using the algorithm described in Section~\ref{sec-equiv}.
Further, we can compare the average-case complexity of an optimal strategy
with the average-case complexities of various ranking strategies, which
can be synthesized for even higher number of coins (more than~$80$). 
In the graph below, we summarize the obtained results.
\bigskip

\noindent
\includegraphics[width=.54\textwidth]{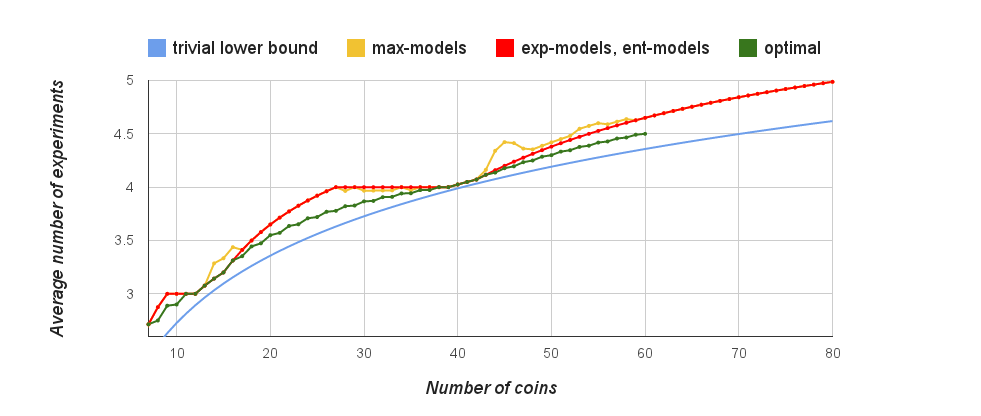}

As the last example, we consider two variants of Mastermind:
MM+col, where we can also ask for all pegs colored by a given color,
and MM+pos, where we can also ask for the color of a given peg.
Using \textsc{Cobra}, we can compute the optimal worst/average case complexity
for $2/8$, $3/6$, and $4/4$ pegs/colors. The results are summarized below.
\begin{center}
{\scriptsize%
\begin{tabulary}{.5\textwidth}{|R|RR|RR|RR|}
\hline
 & \multicolumn{2}{c|}{Mastermind} &
   \multicolumn{2}{c|}{MM+col} &
   \multicolumn{2}{c|}{MM+pos} \\
Size & average      & worst & average      & worst & average      & worst\\\hline
2/8  & 3.67187  & 5     & 3.64062  & 5     & 2        & 2\\
3/6  & 3.19444  & 4     & 3.18981  & 4     & 3        & 3\\
4/4  & 2.78516  & 3     & 2.74609  & 3     & 2.78516  & 3\\\hline
\end{tabulary}}
\end{center}

Let us note that when comparing these results to ``classical'' results
about Mastermind, the following subtle difference in game rules must 
be taken into account: Plays of ``our'' deductive games terminate as soon
as we obtain enough information to reveal the secret code. The ``classical''
Mastermind terminates when the secret code is ``played'', which
may require an extra experiment even if we already know the code. 
Our numbers are valid for the first setup.


\section{Conclusions, Future Work}
\label{sec-concl}

We presented a general framework for modeling and analyzing
deductive games, and we implemented the framework in a software
tool \textsc{Cobra}. Obviously, there are many ways how to improve
the functionality of the presented algorithms and thus extend
the scope of algorithmic analysis to even larger deductive games,
including the ones suggested in bioinformatics 
\cite{Goodrich:MastermindAttack,GET:MastermindPhonotyping-BMC}, or
applied security \cite{FL:MastermindBankPins-ToCS}. Another improvement
may be achieved by tuning the interface to SAT solvers and 
utilizing the sophisticated technology developed in this 
area even more intensively.


\bibliographystyle{abbrv}
\bibliography{deductivegames}

\end{document}